\documentclass{arxiv}

\usepackage{graphicx,amsmath,bm}

\usepackage{setspace}

\usepackage{epstopdf}
\usepackage{color}
\usepackage{tablefootnote}

\usepackage[hyphens]{url}

\usepackage{float}
\usepackage{subcaption}
\usepackage{listings}
\usepackage{mathalfa}
\usepackage{gensymb}
\usepackage{appendix}
\usepackage{hyperref}
\usepackage{amssymb}
\usepackage{multicol}

\begin{document}

\title{Blind Descent: A Prequel to Gradient Descent}


\author{Akshat Gupta\textsuperscript{*}\and Prasad Narahari Raghavendra}
\affilOne{Electrical and Computer Engineering, Carnegie Mellon University\\ \textit{\{akshatgu, praghave\}@andrew.cmu.edu}\\}


\twocolumn[{

\maketitle

\begin{abstract}
We describe an alternative learning method for neural networks, which we call Blind Descent. By design, Blind Descent does not face problems like exploding or vanishing gradients. In Blind Descent, gradients are not used to guide the learning process. In this paper, we present Blind Descent as a more fundamental learning process compared to gradient descent. We also show that gradient descent can be seen as a specific case of the Blind Descent algorithm. We also train two neural network architectures, a multilayer perceptron and a convolutional neural network, using the most general Blind Descent algorithm to demonstrate a proof of concept.
\end{abstract}

\msinfo{}{}{}


}]


\setcounter{page}{1}
\corres
\volnum{1}
\issuenum{1}
\monthyear{May 2020}
\pgfirst{1}
\pglast{100}
\doinum{1}
\articleType{}


\markboth{Akshat et al}{Blind Descent: A Prequel to Gradient Descent}

\section{Introduction}
We see Blind Descent as a prequel to gradient descent. By design, Blind Descent does not require calculation of gradients. Consequently, it does not face certain problems faced by gradient descent, including vanishing and exploding gradients. In gradient descent, there's also a need to store values from forward pass at each node for backpropagation. We do not need to do this for Blind Descent. However, not calculating gradients is what makes this method blind as we are not using gradients to guide the learning proces, hence the name Blind Descent. \\

In later sections, we describe the Blind Descent algorithm and its motivations. We also look at Blind Descent as a generalized learning process, which is more fundamental than gradient descent, which can be seen as a special case of Blind Descent. We also perform experiments on the MNIST dataset to demonstrate the working of the Blind Descent algorithm.

\section{Prior work}
We did not find much prior works to build on. A remotely similar work is on Extreme Learning Machine.\cite{eml} However, the input-hidden layer appears to be randomly initialised and frozen and only the hidden-output layer appears to be minimised using pseudo-inverse solution. Even the incremental learning variant appears to do the same (but with convex optimisation for only one layer).\cite{eml1} Other methods use injected noise.\cite{injectedNoise} But, controlling the noise can be a difficult endeavor and moreover, the data itself can be said to have noise and the required information components. So, we decided to eliminate the noise present in the data instead of introducing additional noise. There are several variants of Extreme Learning Machine (ELM). But, as we can see in the survey, they are either too simplistic and reduce to linear regression or tend towards convex optimisation problems like SVM.\cite{ELMsurvey} However, we did find random optimisation statistical papers of 1960s and one of the most influential papers that directly influenced our research is the backpropagation paper.\cite{backprop}

\section{Blind Descent}
The motivation behind Blind Descent was to try and overcome problems faced when gradients are used to guide the learning process in neural networks. An extreme step in that direction is to eliminate the use of gradients in the learning process. This leads us to Blind Descent. Calculating gradients do have their advantages, as they guide is the direction of steepest descent. Thus, using gradients, we not only know if we're going in the correct direction, but we also know that we're going in the best possible correct direction, locally speaking. Thus at first the idea of doing away with gradients seemed implausible, as we cannot in good conscience do away with the only guiding force we have in the learning process. But once we decided to let go of gradients, we discovered alternate methods that could be used to guide the learning process, which make up the Blind Descent algorithm.\\

In Blind Descent, we do not calculate gradients to guide the learning process. This is a one sentence elevator pitch for the algorithm repeated multiple times in this paper. But then how does the network learn? The weights at each node of the neural network are randomly initialised. Then on every iteration, we update the weights randomly. This randomness although, is guided. Only those updates to weights that lead to a reduction to the overall loss function are kept, other updates are discarded. This is the Blind Descent algorithm in its most basic form. \\

To define the random updates, we need to define the distribution from which the random numbers are picked. The distribution has a mean and a standard deviation, which defines the random updates. We center this distribution on the current value of the weights. The update rule in blind descent is given below:

\[
x^{(t+1)} = \begin{cases} 
x^{(t)} + \mathit{d}(\mu = x^{(t)}, \sigma = \eta) , & \text{if }\mathbf{L}(x^{(t+1)}) < \mathbf{L}(x^{(t)}) \\ 
x^{(t)} & \text{otherwise} \end{cases}
\]\

Here, $x^{(t)}$ is the weight at a node of the neural network at a certain time $t$. $\eta$ is the learning rate, which here defines the standard deviation of the  underlying distribution. $\mathit{d}(\mu, \sigma)$ is a distribution governing the random updates, centered around the value of the weight at the current time step. We call this distribution the \textit{update distribution}. $\mathbf{L}$ is the loss function for the neural network under consideration.\\

The above definition of Blind Descent can have various variants. This is discussed in detail in section \ref{variants}. Different distributions can be used to update the weights. The standard deviations can become smaller and smaller as the loss decreases and we reach closer to a good solution, thus making standard deviation a variable quantity. This would be analogous to a learning schedule used while training neural networks using gradient descent frequently. We can also use concepts of momentum in Blind Descent, thus favoring the direction of previous motion in Blind Descent. \\

Blind Descent, as described above, can also be seen as a generalization to gradient descent. To obtain the equation of gradient descent from Blind Descent, we replace the \textit{update distribution} $\mathit{d}(\mu, \sigma)$ with a deterministic function which is nothing but the gradient of the loss function. This observation shows that Blind Descent is a more fundamental learning process in neural networks compared to gradient descent. Gradient descent can be seen as a specific case of Blind Descent, with deterministic updates.

\section{Experiments with Blind Descent}
In this section we train a neural network using Blind Descent. We train a multilayer perceptron (MLP) and a convolutional neural network (CNN) on the MNIST dataset. We do not assume MNIST to be a standard dataset by any means, but it is a non-trivial dataset with 10 classes.

\subsection{Training MLP with Blind Descent}
 We were able to achieve accuracies close to 70\% on the MNIST dataset. The experiments performed are not optimum and various other techniques like momentum, standard deviation scheduler can be used to improve the results. Our focus is to present a proof of concept showing that Blind Descent works.\\

 For training, we use a two layer deep neural network. The network architecture is [784, 256, 10], where 784 is the input dimensions of MNIST, the first layer has 256 neurons and the output layer has 10 neurons. We use cross entropy loss to measure the loss. While training, we do batch updates. Here are the steps followed during batch updates in Blind descent:
 
\begin{enumerate}
	\item Calculate loss with current weights for entire batch
	\item Do weight updates using Blind Descent
	\item Calculate loss with new weights for entire batch
	\item If new loss is less than previous loss, adopt new weights
\end{enumerate}

\begin{figure}
		\begin{subfigure}{0.5\textwidth}
		\centering
		\includegraphics[width=\linewidth]{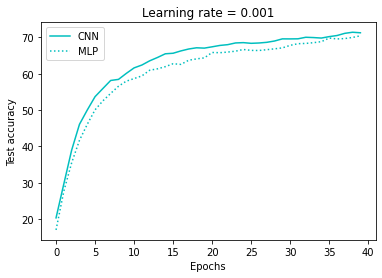} 
		\caption{Accuracy comparison between CNN and MLP architectures for learning rate 0.001 averaged over 10 sample runs.}
		\label{fig:2a}
	\end{subfigure}
	\begin{subfigure}{0.5\textwidth}
		\centering
		\includegraphics[width=\linewidth]{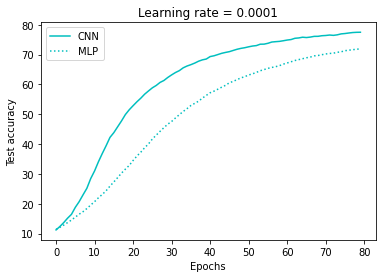}
		\caption{Accuracy comparison between CNN and MLP architectures for learning rate 0.0001 averaged over 10 sample runs.}
		\label{fig:2b}
	\end{subfigure}
	\caption{Blind Descent at work for MLP and CNN architectures}
	\label{fig:1}
\end{figure}

A unique feature we see in Blind Descent is that the training and testing loss keeps increasing along with the increasing accuracy (except the initial drastic decrease in loss). This can be seen in both figure \ref{fig:2}. It might seem unintuitive as we are restricting updates only if the loss on update is lower than the loss before updates. Although this is true for one batch, the loss for the updated parameters for the next batch could be higher for the same parameters. The condition only compared the updated loss to the loss of the current batch. This is the main reason why the loss is increasing (excluding the initial decrease in loss). Even though that happens, the accuracy still increases. Notice that we never put in the condition for the accuracy to increase in our algorithm and it still increases!\\

\subsection{Training CNN with Blind Descent}
\begin{figure}
	\begin{subfigure}{0.5\textwidth}
		\centering
		\includegraphics[width=\linewidth]{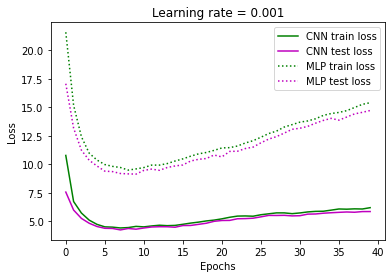} 
		\caption{Training and Test loss for CNN and MLP architectures averaged over 10 sample runs for learning rate = 0.001.}
		\label{fig4a}
	\end{subfigure}
	\begin{subfigure}{0.5\textwidth}
		\centering
		\includegraphics[width=\linewidth]{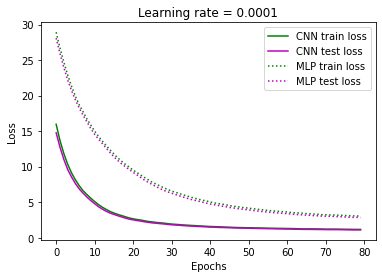}
		\caption{Training and Test loss for CNN and MLP architectures averaged over 10 sample runs for learning rate = 0.0001.}
		\label{fig:4b}
	\end{subfigure}
	\caption{Training and Tess Loss comparison between CNN and MLP architectures.}
	\label{fig:2}
\end{figure}

We demonstrate that even a single CNN layer can perform well with blind descent and achieve accuracy of about 73\% even when we have not used momentum, learning rate schedule, automated hyper-parameter tuning etc. We choose a CNN architecture having one CNN layer consisting of 16 kernels, each of size $3\times 3$. We have a max pool layer after that with kernel size = 3, stride = 2 and padding = 1. The output of the max pooling layer is connected to a linear layer consisting of 128 neurons. The results for the CNN architecture are shown in figure \ref{fig:1} and \ref{fig:2}.

\section{Variants}\label{variants}
This section highlights many variants of gradient descent which can help in faster Blind Descent. 
\begin{itemize}
    \item \textit{Update Distributions}: Different \textit{update distributions} can be used for the learning process using Blind Descent. In our work, we have used a normal distribution. Update distributions can in principle be stochastic functions and have both deterministic and non-deterministic components. 

    \item \textit{Standard Deviation Scheduling}: We can schedule the standard deviation based on the number of successful updates in Blind descent. As the loss decreases, it is likely that with a larger standard deviation, we're likely to miss an optimum update that reduces the loss. To counter that, we can reduce the standard deviation based as the number of successful updates decreases. We can also use the test accuracy for standard deviation scheduling. This is analogous to a learning rate schedule used with gradient descent.\cite{lrSchedule}
    
    \item \textit{Momemtum}: We can use the concept of momentum in Blind Descent as well (similar to gradient descent\cite{momentum}). In the version of Blind Descent presented in this paper, the mean of the \textit{update distribution} is at the current value of the weight. We can build up momentum with successful updates to the weights and shift the mean of the \textit{update distribution} in direction of the building momentum. This moves the mean of the \textit{update distribution} away from the current value of the weights and towards the value suggested by the building momentum, thus accelerating the learning process.
    
    \item \textit{Combining other learning techniques}: We can use Blind Descent with many other training techniques. Layer by layer training, first suggested in \cite{cascade} can be used with Blind Descent. We can also incorporate Blind Descent while transfer learning. It is also possible to train a Blind Descent Network further using gradient descent and use Blind Descent training as an initialization.
    
\end{itemize}

\section{Discussion}
In this paper, we introduce the Blind Descent algorithm as a generalization to gradient descent. We trained two neural network architectures, namely a multilayer perceptron and a convolutional neural network to show that the algorithm works. Blind Descent has various advantages over gradient descent including not having to save forward pass values and calculating gradients at each node. This happens by design as we are not calculating gradients. Accordingly, Blind Descent does not suffer with limitations of gradient descent like vanishing and exploding gradients. \\

In Blind Descent we can choose to accept or reject an update, such a luxury is not available in case of gradient descent. This also remedies another disadvantage of gradient descent. For an instance of gradient descent, we are bound by the network initialization and update steps taken in the past, which are locally optimal but might be sub-optimal globally. Blind descent on the other hand gives us the opportunity to explore a larger solution space at every step, thus giving us a possible way out of a sub-optimal solution. We believe this feature of Blind Descent can be used to augment gradient descent, allowing us to explore a larger solution space.\\

Even though Blind Descent has some merits, without gradients, we do not know the direction in which we should proceed and thus we are blind. Thus we do not expect Blind Descent to compete with Gradient Descent in terms of performance.\\

Future work with Blind Descent includes implementing the many variants of Blind Descent as discussed in section \ref{variants}. Larger neural networks and other complex architectures can also be trained to test the Blind Descent algorithm. We also have a whole spectrum of \textit{update distributions} to explore. \\




\begin{thebibliography}{1}

\bibitem{eml}
Guang-Bin Huang, Qin-Yu Zhu, Chee-Kheong Siew, \textit{Extreme learning machine: Theory and applications}, vol. 70, pp. 489 - 501, Neurocomputing, Elsevier, 2005

\bibitem{eml1}
Guang-Bin Huang, Qin-Yu Zhu, Chee-Kheong Siew, \textit{Universal Approximation Using Incremental Constructive Feedforward Networks With Random Hidden Nodes}, IEEE Transactions on Neural Networks, Vol. 17, 2006.

\bibitem{injectedNoise}
J. L. Maryak,  D. C. Chin, \textit{Global random optimization by simultaneous perturbation stochastic approximation}, Proceeding of the Winter Simulation Conference IEEE, 2001

\bibitem{ELMsurvey}
Guang-Bin Huang, Dian Hui Wang, Yuan Lan, \textit{Extreme learning machines: a survey}, International Journal of Machine Learning and Cybernetics, Vol. 2, pp. 107–12, 2011

\bibitem{backprop}
David E. Rumelhart, Geoffrey E. Hinton, Ronald J. Williams, \textit{Learning representations by back-propagating errors}, Vol. 323, Nature, 1986

\bibitem{lrSchedule}
Schaul, Tom, Sixin Zhang, and Yann LeCun. \textit{No more pesky learning rates.}, In International Conference on Machine Learning, pp. 343-351. 2013.

\bibitem{momentum}
Qian, Ning. \textit{On the momentum term in gradient descent learning algorithms.}, Neural networks 12, no. 1 (1999): 145-151.

\bibitem{cascade}
Fahlman, Scott E., and Christian Lebiere. \textit{The cascade-correlation learning architecture.} Advances in neural information processing systems. 1990.

\end{thebibliography}
\end{document}